\documentclass{article}

\usepackage{arxiv}

\usepackage[utf8]{inputenc} 
\usepackage[T1]{fontenc}    
\usepackage{hyperref}       
\usepackage{url}            
\usepackage{booktabs}       
\usepackage{amsmath, amssymb}
\usepackage{amsfonts}       
\usepackage{nicefrac}       
\usepackage{microtype}      
\usepackage{lipsum}
\usepackage{amscd}
\usepackage{multicol}
\usepackage{graphicx}
\usepackage[T1]{fontenc}
\usepackage[utf8]{inputenc}
\usepackage{authblk}

\newtheorem{theorem}{Theorem}[section]

\newtheorem{definition}[theorem]{Definition}
\newtheorem{problem}[theorem]{Problem}

\title{Mode Collapse and Regularity of Optimal Transportation Maps}
\author[1]{Na Lei}
\author[2]{Yang Guo}
\author[2]{Dongsheng An}
\author[2]{Xin Qi}
\author[1]{Zhongxuan Luo}
\author[3]{Shing-Tung Yau}
\author[2]{Xianfeng Gu}
\affil[1]{School of Software and Technology, Dalian University of Technology}
\affil[2]{Department of Computer Science, Stony Brook University}
\affil[3]{Department of Mathematics, Harvard University}

\begin{document}
\maketitle
\begin{abstract}%
 This work builds the connection between the regularity theory of optimal transportation map, Monge-Amp\`{e}re equation and GANs, which gives a theoretic understanding of the major drawbacks of GANs: convergence difficulty and mode collapse.

According to the regularity theory of Monge-Amp\`{e}re equation, if the support of the target measure is disconnected or just non-convex, the optimal transportation mapping is discontinuous. General DNNs can only approximate continuous mappings. This intrinsic conflict leads to the convergence difficulty and mode collapse in GANs.

We test our hypothesis that the supports of real data distribution are in general non-convex, therefore the discontinuity is unavoidable using an Autoencoder combined with discrete optimal transportation map (AE-OT framework) on the CelebA data set. The testing result is positive. Furthermore, we propose to approximate the continuous Brenier potential directly based on discrete Brenier theory to tackle mode collapse. Comparing with existing method, this method is more accurate and effective.
\end{abstract}


\section{Introduction}
Generative Adversarial Networks (GANs, \cite{goodfellow2014generative}) emerge as one of the dominant approaches for unconditional image generating. GANs have successfully shown their amazing capability of generating realistic looking and visual pleasing images. Typically, a GAN model consists of an unconditional generator that regresses real images from random noises and a discriminator that measures the difference between generated samples and real images. Despite GANs' advantages, they have critical drawbacks. 1) Training of GANs are tricky and sensitive to hyperparameters. 2) GANs suffer from mode collapsing. Recently Meschede et. al (\cite{mescheder2018training}) studied $9$ different GAN models and variants showing that gradient descent based GAN optimization is not always convergent. \textbf{The goal of this work is to improve the theoretic understanding of these difficulties and propose methods to tackle them fundamentally.}

\begin{figure}[tp]
\centering
\begin{tabular}{c}
\includegraphics[width=0.7\textwidth]{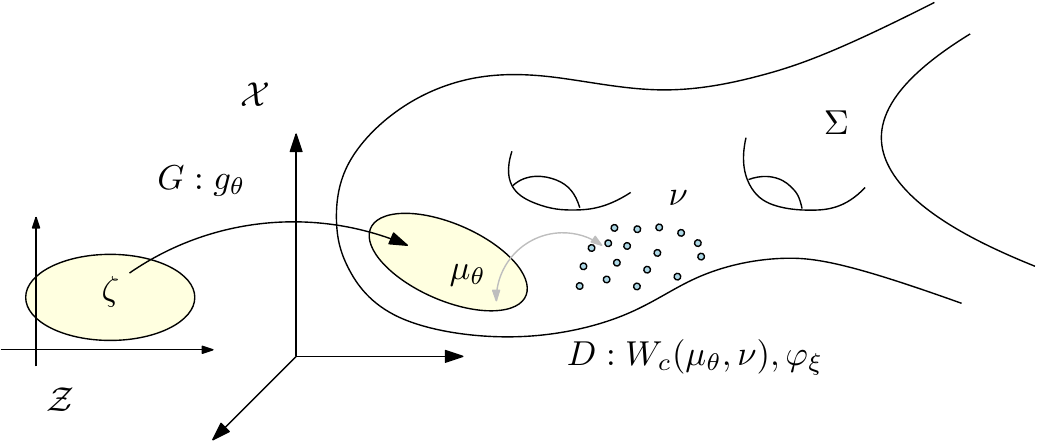}
\end{tabular}
\caption{Wasserstein GAN framwork. \label{fig:GAN_framework}}
\end{figure}

\vspace{2mm}

\noindent\textbf{Optimal Transportation View of GANs} 
Recent promising successes are making GANs more and more attractive (e.g. \cite{radford2015unsupervised,sonderby2016amortised,zhu2017unpaired}). Among various improvements of GANs, one breakthrough has been made by incorporating GANs with optimal transportation (OT) theory (\cite{villani2008optimal}), such as in works of WGAN (\cite{arjovsky2017wasserstein}), WGAN-GP~(\cite{gulrajani2017improved}) and RWGAN~(\cite{guo2017relaxed}). In WGAN framework, the generator computes the optimal transportation map from the white noise to the data distribution, the discriminator computes the Wasserstein distance between the real and the generated data distributions.

Fig.~\ref{fig:GAN_framework} illustrates the framework of WGAN. The image space is $\mathcal{X}$, the real data distribution $\nu$ is concentrated on a manifold $\Sigma$ embedded in $\mathcal{X}$. $\mathcal{Z}$ is the latent space, $\zeta$ is the white noise (Gaussian distribution). The generator computes a transformation map $g_\theta$, which maps $(\mathcal{Z},\zeta)$ to $(\mathcal{X},\mu_\theta)$; the discriminator computes the Wasserstein distance between $\mu_\theta$ and the real distribution $\nu$ by finding the Kontarovich potential $\varphi_\xi$ (refer to Eqn. \ref{eqn:DP2}).

In principle, the GAN model accomplishes two major tasks: 1) manifold learning, discovering the manifold structure of the data; 2) probability transformation, transforming a white noise to the data distribution. Accordingly, the generator map $g_\theta:(\mathcal{Z},\zeta)\to(\Sigma,\mu_\theta)$ can be further decomposed into two steps,
\begin{equation}
g_\theta:
\begin{CD}
(\mathcal{Z},\zeta)   \xrightarrow{T}  
 (\mathcal{Z},\mu) \xrightarrow{g}  (\Sigma, \mu_\theta)
\end{CD}
\label{eqn:generator_map_decomposition}
\end{equation}
where $T$ is a transportation map, maps the white noise $\zeta$ to $\mu$ in the latent space $\mathcal{Z}$, $g$ is the manifold parameterization, maps local coordinates in the latent space to the manifold $\Sigma$. Namely, $g$ gives a local chart of the data manifold $\Sigma$, $T$ realizes the probability measure transformation. The goal of the GAN model is to find $g_\theta$, such that the generated distribution $\mu_\theta$ fits the real data distribution $\nu$, namely
\begin{equation}
(g_\theta)_\# \zeta = \nu.
\label{eqn:goal_of_GAN}
\end{equation}

\noindent\textbf{Regularity Analysis for mode collapse} By manifold structure assumption, the local chart representation $g: \mathcal{Z}\to \Sigma$ is continuous. Unfortunately, the continuity of the transportation map $T:\zeta\to \mu$ can not be guaranteed. Even worse, according to the regularity theory of optimal transportation map, except very rare situations, the transportation map $T$ is always discontinuous. In more details, unless the support of $\mu$ is convex, there are non-empty singularity sets in the domain of $T$, where $T$ is discontinuous. By Eqn.~\ref{eqn:generator_map_decomposition} and \ref{eqn:goal_of_GAN}, $\mu=(g^{-1})_\#\nu$ is determined by the real data distribution $\nu$ and the encoding map $g^{-1}$, it is highly unlikely that the support of $\mu$ is convex.

On the other hand, the deep neural networks (DNNs) can only model continuous mappings. For example, the commonly used ReLU DNNs can only represent piece-wise linear mappings. But the desired mapping  itself is discontinuous. This intrinsic conflict explains the fundamental difficulties of GANs: \textbf{Current GANs search a discontinuous mapping in the space of continuous mappings, the searching will not converge or converge to one continuous branch of the target mapping, leading to a mode collapse.}

\vspace{2mm}
\noindent\textbf{Solution to mode collapse} 
We propose a solution to the mode collapse problem based on the Brenier theory of optimal transportation (\cite{villani2008optimal}). According to Brenier theorem \ref{thm:Brenier}, under the quadratic distance cost function, the optimal transportation map is the gradient of a convex function, the so-called Brenier potential. Under mild conditions, the Brenier potential is always continuous and can be represented by DNNs. \textbf{We propose to find the continuous Brenier potential instead of the discontinuous transportation map.}

\vspace{2mm}
\noindent\textbf{Contributions}
This work improves the theoretic understanding of the convergence difficulty and mode collapse of GANs from the perspective of optimal transportation; builds connections between the regularity theory of optimal transportation map, Monge-Amp\`{e}re equation, and GANs; proposes solutions to conquer mode collapse based on Brenier theory.

This paper is organized as follows: in Section \ref{sec:theory}, we briefly introduce the theory of optimal transportation; in Section \ref{sec:alg}, we give a computational algorithm based on the discrete version of Brenier theory; in Section \ref{sec:gan}, we explain the mode collapse issue using Monge-Amp\`{e}re regularity theory, and propose a novel method. Furthermore, we test our hypothesis that general transportation maps in GANs are discontinuous with proposed method. The testing results are reported in Section \ref{sec:gan} as well. Finally, we draw the conclusion in Section \ref{sec:conclusion}.

\section{Previous Work}

\textbf{Generative adversarial networks.} Generative adversarial networks (GANs) are technique for training generative models to produce realistic examples from an unknown distribution (\cite{goodfellow2014generative}). In particular, the GAN model consists of a generator network that maps latent vectors, typically drawn from a standard Gaussian distribution, into real data distribution and a discriminator network that aims to distinguish generated data distribution with the real one. Training of GANs is unfortunately found to be tricky and one major challenge is called \textit{mode collapse}, which refers to a lack of diversity of generated samples. This commonly happens when trained on multimodal distributions. For example on a dataset that consists of images of ten handwritten digits, generators might fail to generator some digits (\cite{gulrajani2017improved}).
Prior works have observed two types of mode collapse, i.e fail to generate some modes entirely, or only generating a subset of a particular mode (\cite{goodfellow2016nips, tolstikhin2017adagan, arora2017gans, donahue2016adversarial, metz2016unrolled, reed2016generative}). Several explanatory hypothesis to mode collapse have been made, including proper objective functions (\cite{arjovsky2017wasserstein,arora2017generalization}) and weak discriminators (\cite{metz2016unrolled,salimans2016improved,arora2017generalization,li2017towards}). 

Three main approaches to mitigate mode collapse include employing inference networks in addition to generators (e.g. \cite{donahue2016adversarial, dumoulin2016adversarially, srivastava2017veegan}), discriminator augmentation (e.g. \cite{che2016mode,salimans2016improved,karras2017progressive,lin2018pacgan}) and improving optimization procedure during GAN training (e.g. \cite{metz2016unrolled}). However these methods measure the difference between implicit distributions by a neural network (i.e discriminator), whose training relies on solving a non-convex optimization problem that might lead to non-convergence (\cite{li2017towards,mescheder2018training}). In contrast, the proposed method metrics the distance between two distributions by $L^2$ Wasserstein distance, which can be computed under a convex optimization framework. Furthermore the optimal solution provides a transport map transforming the source distribution to the target distribution, and essentially serves as a generator in the feature space. Empirically, the distribution generated by this generator is theoretically guaranteed to be identical to the real one, without mode collapse or artificial mode invention. 

\noindent\textbf{Optimal Transport.}
Optimal transport problem attracted the researchers attentions since it was proposed in 1940s, and there were vast amounts of literature in various kinds of fields like computer vision and natural language processing. We recommend the readers to refer to \cite{peyre2018COT} and \cite{solomon2018OTDD} for detailed information.

Under discrete optimal transport, in which both the input and output domains are Dirac masses, we can use standard linear programming (LP) to model the problem. To facilitate the computation complexity of LP,  \cite{cuturi2013sinkhorn} added an entropic regularizer into the original problem and it can be quickly
computed with the Sinkhorn algorithm. By the introduction of fast convolution, Solomon \& Guibas (\cite{Solomoon2015ConvWD}) improved the computational efficiency of the above algorithm. The smaller the coefficient of the regularizer, the solution of the regularized problem is closer to the original problem. However, when the coefficient is too small, the sinkhorn algorithm cannot find a good solution. Thus, it can only approximate the original problem coarsely.

When computing the transport map between continuous and point-wise measures, i.e. the semi-discrete optimal transport, (\cite{gu2013theory}) proposed to minimize a convex energy through the connection between the OT problem and convex geometry. Then by the link between c-transform and Legendre dual theory, the authors of \cite{lei2017geometric} found the equivalence between solution of the Kantorovich duality problem and that of the convex energy minimization.

If both the input and output are continuous densities, the OT problem can be treated to solve the Monge-Amp{\'e}re equation. (\cite{Benamou2002CFM,Benamou2014NumSolution,Papadakis2014ProSplit}) solved this PDE by computational fluid dynamics with an additional virtual time dimension. But this kind of problem is both time consuming and hard to extend to high dimensions.

\section{Optimal Mass Transport theory}
\label{sec:theory}

In this subsection, we will introduce basic concepts and theorems in classic optimal transport theory, focusing on Brenier's approach, and their generalization to the discrete setting. Details can be found in \cite{villani2008optimal}.

\noindent\textbf{Optimal Transportation Problem} 
Suppose $X\subset \mathbb{R}^d$, $Y\subset \mathbb{R}^d$  are two subsets of $d$-dimensional Euclidean space $\mathbb{R}^d$, $\mu, \nu$ are two probability measures defined on $X$ and $Y$ respectively, with density functions
$
\mu(x)= f(x)dx, \nu(y)=g(y)dy.
$
Suppose the total measures are equal, $\mu(X)=\nu(Y)$, namely
$
    \int_X f(x)dx = \int_Y g(y)dy.
$
We only consider maps which preserve the measure.
\begin{definition}[Measure-Preserving Map]
\label{def:omt}
A map $T: X\to Y$ is \emph{measure preserving} if for any measurable set $B\subset Y$, the set $T^{-1}(B)$ is $\mu$-measurable and $\mu(T^{-1}(B))=\nu(B)$, i.e.
$
    \int_{T^{-1}(B)} f(x)dx = \int_B g(y) dy.
$
Measure-preserving condition is denoted as $T_\#\mu = \nu$, where $T_\#\mu$ is the push forward measure induced by $T$.
\end{definition}

Given a cost function $c(x,y):X\times Y\to \mathbb{R}_{\ge 0}$, which indicates the cost of moving each unit mass from the source to the target,
the total {\it transport cost} of the map $T:X\to Y$ is defined to be $\int_X c(x,T(x)) d\mu(x)$.

The Monge's problem of optimal transport arises from finding the measure-preserving map that minimizes the total transport cost.
\begin{problem}
\label{prob: Monge}
[Monge's Optimal Mass Transport (MP)  (\cite{bonnotte2013knothe})] Given a transport cost function $c: X\times Y\to \mathbb{R}$, find the measure preserving map $T:X\to Y$ that minimizes the total transport cost
\begin{equation}
(MP)\hspace{6mm}  \min_{T_{\#}\mu=\nu} \int_X c(x,T(x)) d\mu(x).
    \label{eq:MP}
\end{equation}
\end{problem}
\begin{definition}[Optimal Transportation Map]
The solutions to the Monge's problem is called the \emph{optimal transportation map}, whose  total transportation cost is called the \emph{Wasserstein distance} between $\mu$ and $\nu$, denoted as $\mathcal{W}_c(\mu,\nu)$.
\begin{equation}
    \mathcal{W}_{c}(\mu,\nu) = \min_{T_\#\mu=\nu} \int_X c(x,T(x)) d\mu(x).
\end{equation}
\end{definition}
For the cost function being the $L^1$ norm, Kontarovich relaxed transportation maps to transportation plans, and proposed linear programming method to solve this problem. We introduce the details of Kontarovich's approach in Appendix \ref{sec:Kontarovichmethod}.

\vspace{2mm}
\noindent\textbf{Brenier's Approach} For quadratic Euclidean distance cost, the existence, uniqueness and the intrinsic structure of the optimal transportation map were proven by Brenier (\cite{brenier1991polar}).
 \begin{theorem}[\cite{brenier1991polar}]
 \label{thm:Brenier}
 Suppose $X$ and $Y$ are the Euclidean space $\mathbb{R}^d$ and the transportation cost is the quadratic Euclidean distance
  $c(x,y) = 1/2\|x-y\|^2$. Furthermore $\mu$ is absolutely continuous and $\mu$ and $\nu$ have finite second order moments,
 $
    \int_X |x|^2 d\mu(x) + \int_Y |y|^2 d\nu(y) < \infty,
 $
  then there exists a convex function $u: X \to \mathbb{R}$, the so-called Briener potential, its gradient map $\nabla u$ gives the solution to the Monge's problem,
  \begin{equation}
    (\nabla u)_{\#} \mu = \nu.
  \end{equation}
  The Brenier potential is unique upto a constant, hence the optimal transportation map is unique.
 \end{theorem}
Assume the Briener potential is $C^2$ smooth, the it is the solution to the following Monge-Am\`{e}re equation:
\begin{equation}
\text{det} \left(\frac{\partial^2 u}{\partial x_i \partial x_j}\right) (x) = \frac{\mu(x)}{\nu\circ \nabla u(x)}
\end{equation}
In general cases, the Brenier potential $u$ satisfing the transport condition
$(\nabla u)_{\#} \mu = \nu$ can be seen as a weak form of Monge-Amp\`{e}re equation, coupled with the boundary condition $\nabla u(X)=Y$. Hence $u$ is called a \emph{Brenier solution}.

\begin{definition}[Legendre Transform] Given a function $\varphi:\mathbb{R}^n\to \mathbb{R}$, its Legendre transform is defined as
\begin{equation}
    \varphi^*(y):=\sup_x \left(\langle x, y \rangle - \varphi(x)\right).
    \label{eqn:Legendre_transform}
\end{equation}
\end{definition}

\begin{figure}[t]
\centering
\begin{tabular}{c}
\includegraphics[width=0.65\textwidth]{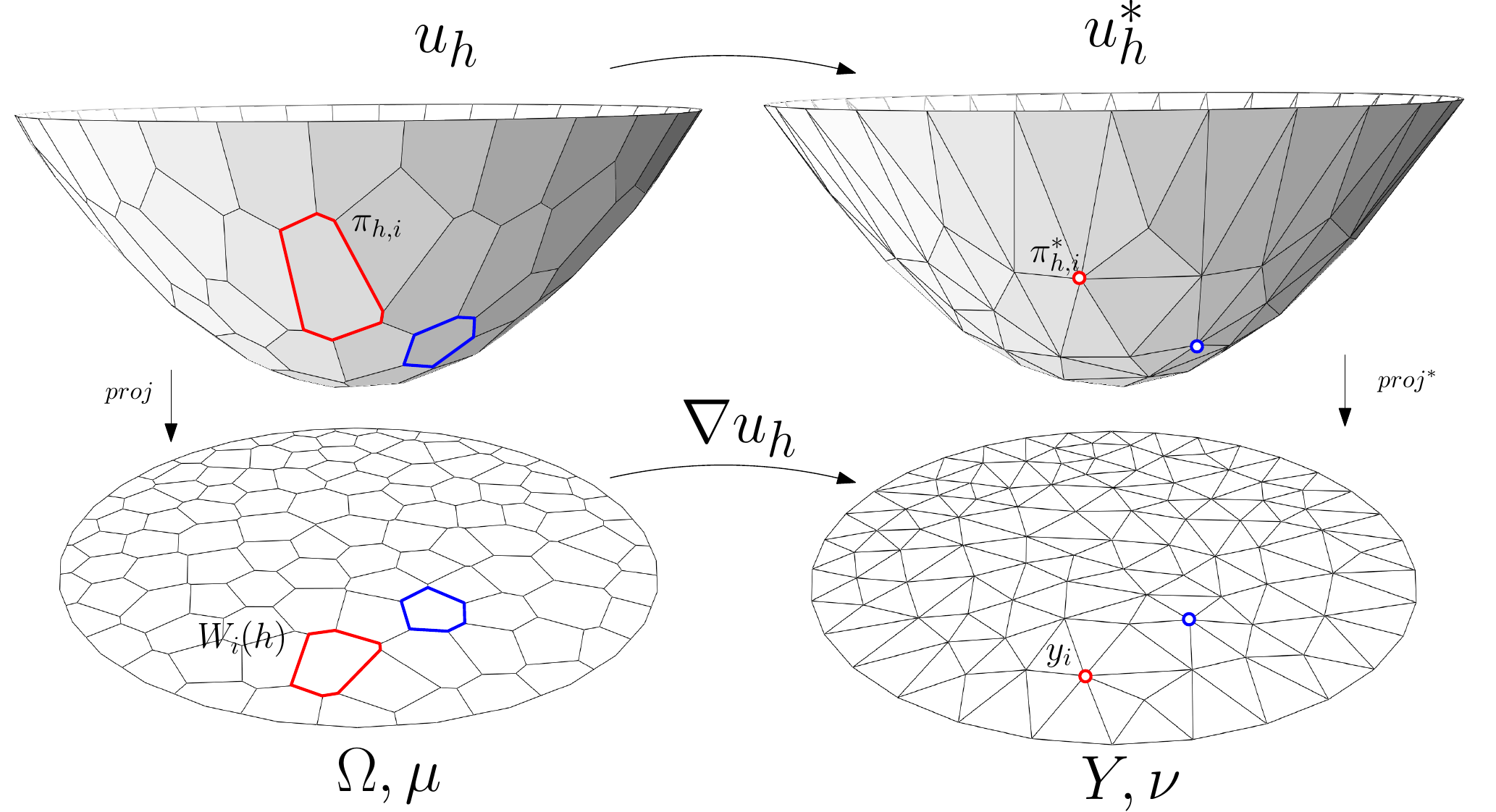}
\end{tabular}
\caption{PL Brenier potential (left) and its Legendre dual (right). \label{fig:PL_Brenier_potential}}
\end{figure}

In practice, the Brenier solution is approximated by the so-called Alexandrov solution.
\begin{definition}[Sub-gradient]
Let $u:\mathbb{R}^d\to\mathbb{R}$ be a convex function. Its \emph{sub-gradient or sub-differential} at a point $x$ is defined as
\begin{equation}
    \partial u(x) := \{ y\in \mathbb{R}^d | u(z)\ge u(x) + \langle y, z-x\rangle , \forall z\in \mathbb{R}^d\}.
\end{equation}
\end{definition}
A convex function is Lipschitz, hence it is differentiable almost everywhere.
\begin{definition}[Alexandrov Solution]
If a convex function $v:\mathbb{R}^d\to\mathbb{R}$ satisfies
\begin{equation}
    |\partial v(E)| = \int_E \frac{f(x)}{g\circ \nabla v(x)} dx,~~~~\forall E\subset X
\end{equation}
then we say $v$ is an Alexandrov solution to the Monge-Amp\`{e}re equation.
\end{definition}

\noindent\textbf{Regularity of Optimal Transportation Maps}
Let $\Omega$ and $\Lambda$ be two bounded smooth open sets in $\mathbb{R}^d$, let $\mu=fdx$ and $\nu=gdy$ be two probability measures on $\mathbb{R}^d$ such that $f|_{\mathbb{R}^d\setminus \Omega}=0$ and $g|_{\mathbb{R}^d\setminus \Lambda}=0$. Assume that $f$ and $g$ are bounded away from zero and infinity on $\Omega$ and $\Lambda$, respectively,

According to Caffarelli (\cite{caffarelli1992regularity}), if $\Lambda$ is convex, then the Alexandrov solution $u$ is strictly convex, furthermore
\begin{enumerate}
    \item If $\lambda \le f$, $g\le 1/\lambda$ for some $\lambda > 0$, then $u\in C_{loc}^{1,\alpha}(\Omega)$.
    \item If $f\in C_{loc}^{k,\alpha}(\Omega)$ and $g\in C_{loc}^{k,\alpha}(\Lambda)$, with $f,g>0$, then $u\in C_{loc}^{k+2,\alpha}(\Omega)$, $(k\ge 0, \alpha\in (0,1))$
\end{enumerate}
Here $C_{loc}^{k,\alpha}$ represents the $k$-th order H\"{o}lder continuous with exponant $\alpha$ function space. If $\Lambda$ is not convex, there exist $f$ and $g$ smooth such that $u\not\in C^1(\Omega)$, the optimal transportation map $\nabla u$ is discontinuous at singularities. $u$ is differentiable if its subgradient $\partial u$ is a singleton. We classify the points according to the dimensions of their subgradients, and define the sets
\[
    \Sigma_k(u) := \left\{ x\in \mathbb{R}^d| \text{dim}(\partial u(x))= k \right\}, \quad k=0,1,2\dots, d.
\]
It is obvious that $\Sigma_0(u)$ is the set of regular points, $\Sigma_k(u)$, $k>0$ are the set of singular points. We also define the \emph{reachable subgradients} at $x$ as
\[
    \nabla_{*} u(x):= \left\{ \lim_{k\to \infty } \nabla u(x_k) | x_k \in \Sigma_0, x_k \to x \right\}.
\]
It is well known that the subgradient equals to the convex hull of the reachable subgradient,
\[
    \partial u(x) = \text{Convex Hull}(\nabla_* u(x)).
\]

\begin{theorem}[Regularity] Let $\Omega,\Lambda\subset\mathbb{R}^d$ be two bounded open sets, let $f,g:\mathbb{R}^d\to \mathbb{R}^+$ be two probability densities, that are zero outside $\Omega$, $\Lambda$ and are bounded away from zero and
infinity on $\Omega$, $\Lambda$, respectively. Denote by $T = \nabla u : \Omega \to \Lambda$ the optimal transport map provided by theorem \ref{thm:Brenier}. Then there exist two relatively closed sets $\Sigma_\Omega \subset \Omega$ and
$\Sigma_\Lambda \subset \Lambda$ with $|\Sigma_\Omega| = |\Sigma_\Lambda| = 0$ such that $T : \Omega \setminus \Sigma_\Omega \to \Lambda \setminus \Sigma_\Lambda$ is a homeomorphism of
class $C^{0,\alpha}_{loc}$ for some $\alpha>0$.
\label{thm:regularity}
\end{theorem}

\begin{figure}[t]
\centering
\begin{tabular}{c}
\includegraphics[width=0.75\textwidth]{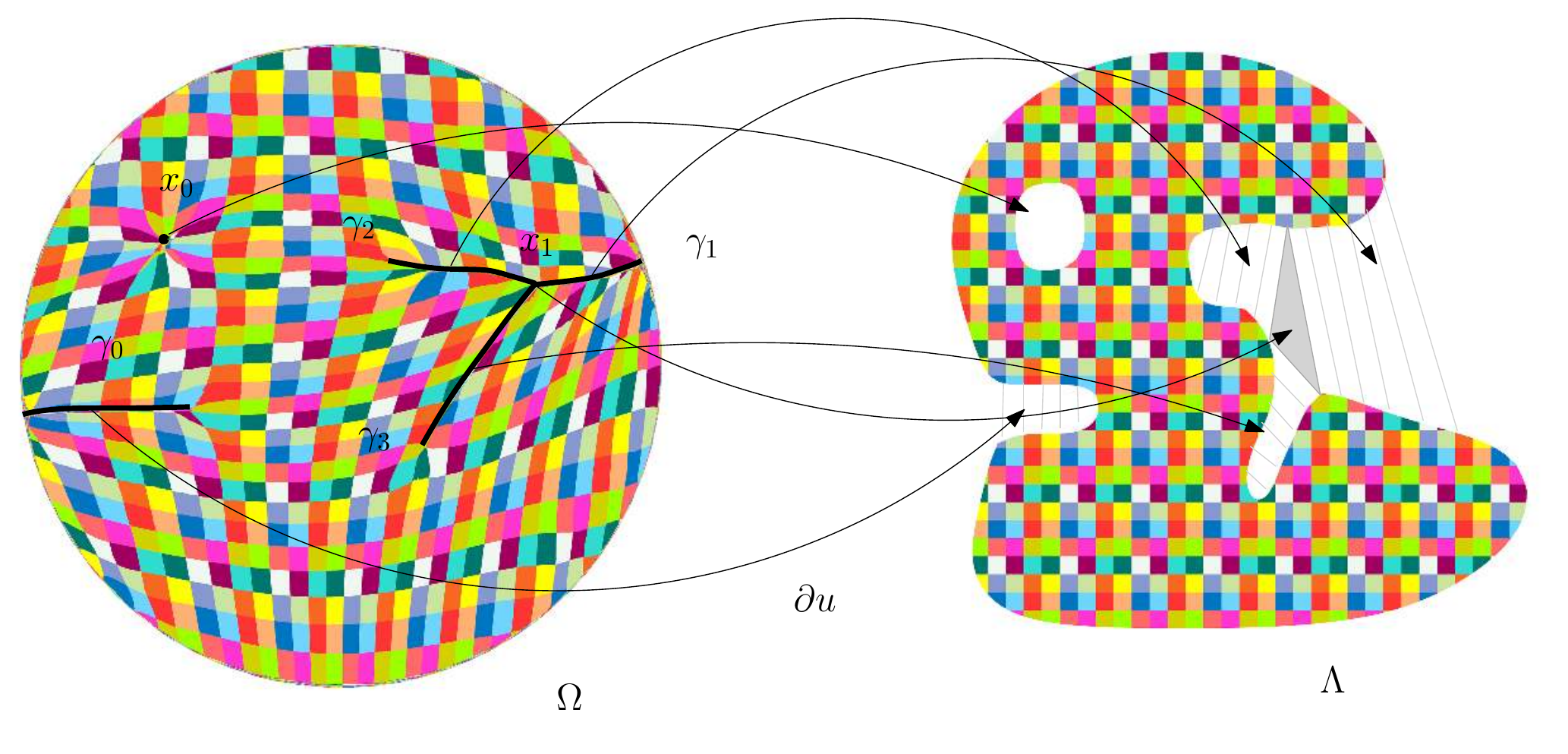}
\end{tabular}
\vspace{-3mm}
\caption{Singularity structure of an optimal transportation map. \label{fig:singularity}}
\end{figure}

Fig.~\ref{fig:singularity} illustrates the singularity set structure of an optimal transportation map $\nabla u:\Omega\to \Lambda$, computed using the algorithm based on theorem \ref{thm:gu_yau}. We obtain
\[
     \Sigma_0 = \Omega \setminus \{\Sigma_1\cup \Sigma_2\},~~\Sigma_1 = \bigcup_{k=0}^3 \gamma_k, ~~\Sigma_2 = \{ x_0, x_1\}.
\]
The subgradient of $x_0$, $\partial u(x_0)$ is the entire inner hole of $\Lambda$, $\partial u(x_1)$ is the shaded triangle. For each point on $\gamma_k(t)$, $\partial u(\gamma_k(t))$ is a line segment outside $\Lambda$. $x_1$ is the bifurcation point of $\gamma_1,\gamma_2$ and $\gamma_3$. The Brenier potential on $\Sigma_1$ and $\Sigma_2$ is not differentiable, the optimal transportation map $\nabla u$ on them are discontinuous.

\section{Discrete Brenier Theory}
\label{sec:alg}

Brenier's theorem can be directly generalized to the discrete situation. In GAN models, the source measure $\mu$ is given as a uniform (or Gaussian) distribution defined on a convex compact domain $\Omega$; the target measure $\nu$ is represented as the empirical measures, which is the sum of Dirac measures
$
    \nu = \sum_{i=1}^n \nu_i \delta(y-y_i),
    \label{eqn:discrete_target_measure}
$
where $Y=\{y_1,y_2,\cdots,y_n\}$ are training samples, weights $\sum_{i=1}^n \nu_i = \mu(\Omega)$. Each training sample $y_i$ corresponds to a supporting plane of the Brenier potential, denoted as
\begin{equation}
    \pi_{h,i}(x):=\langle x,y_i \rangle +h_i,
\end{equation}
where the height $h_i$ is a variable. We represent all the height variables as $h=(h_1,h_2,\cdots,h_n)$.

An \emph{envelope} of a family of hyper-planes in the Euclidean space is a hyper-surface that is tangent to each member of the family at some point, and these points of tangency together form the whole envelope. As shown in Fig.~\ref{fig:PL_Brenier_potential}, the Brenier potential $ u_h:\Omega\to \mathbb{R}$ is a piecewise linear convex function determined by $h$, which is the upper envelope of all its supporting planes,
\begin{equation}
    u_h(x) = \max_{i=1}^n \{\pi_{h,i}(x)\} = \max_{i=1}^n \left\{\langle x, y_i\rangle + h_i\right\}.
    \label{eqn:PL_Brenier_potential}
\end{equation}
The graph of Brenier potential is a convex polytope. Each supporting plane $\pi_{h,i}$ corresponds to a facet of the polytope. The projection of the polytope induces a cell decomposition of $\Omega$, each supporting plane $\pi_i(x)$ projects onto a cell $W_i(h)$,
\begin{equation}
    \Omega = \bigcup_{i=1}^n W_i(h) \cap\Omega, \quad W_i(h) := \{ p \in \mathbb{R}^d | \nabla u_h(p) = y_i \}.
    \label{eqn:cell}
\end{equation}
the cell decomposition is a \emph{power diagram}.

The $\mu$-measure of $W_i\cap\Omega$ is denoted as 
$
    w_i(h) := \mu(W_i(h)) = \int_{W_i(h)\cap \Omega} d\mu.
$
The gradient map $\nabla u_h:\Omega\to Y$ maps each cell $W_i(h)$ to a single point $y_i$,
$
    \nabla u_h : W_i(h)\mapsto y_i, i = 1,2,\dots, n.
$

Given the target measure $\nu$ in Eqn.~\ref{eqn:discrete_target_measure}, there exists a discrete Brenier potential in Eqn.~\ref{eqn:PL_Brenier_potential}, whose projected $\mu$-volume of each facet $w_i(h)$ equals to the given target measure $\nu_i$. This was proved by Alexandrov in convex geometry.

\begin{theorem} [\cite{alex}]
Suppose $\Omega$ is a compact convex polytope with non-empty
interior in $\mathbb{R}^n$, $n_1, ..., n_k \subset \mathbb{R}^{n+1}$ are
distinct $k$ unit vectors, the $(n+1)$-th coordinates are negative, and $\nu_1, ..., \nu_k >0$ so that $\sum_{i=1}^k \nu_i=vol(\Omega)$. Then there exists convex polytope $P\subset \mathbb{R}^{n+1}$ with exact $k$ codimension-1 faces$F_1,\dots,F_k$ so that $n_i$ is the normal vector to $F_i$ and the intersection between $\Omega$ and the projection of $F_i$ is with volume $\nu_i$. Furthermore, such $P$ is unique up to vertical translation.
\end{theorem}

Alexandrov's proof for the existence is based on algebraic topology, which is not constructive. Recently, Gu et al. gave a contructive proof based on the variational approach.
 \begin{theorem}[\cite{gu2013theory}]
 \label{thm:gu_yau}
 Let $\mu$ a probability measure defined on a compact convex domain $\Omega$ in $\mathbb{R}^d$, $Y=\{y_1, y_2, \ldots, y_n\}$ be a set of distinct points in $\mathbb{R}^d$. Then for any $\nu_1, \nu_2, \ldots, \nu_n > 0$ with $\sum_{i=1}^n \nu_i = \mu(\Omega)$, there exists $h = (h_1, h_2, \ldots, h_n) \in \mathbb{R}^n$, unique up to adding a constant $(c,c, \ldots, c)$, so that $w_i(h)  = \nu_i$, for all $i$. The vector $h$ is the unique minimum argument of the following convex energy
 \begin{equation}
    \label{eqn:alexandrov_potential}
     E(h) = \int_0^h \sum_{i=1}^n w_i(\eta)d\eta_i - \sum_{i=1}^n h_i\nu_i,
 \end{equation}
 defined on an open convex set
 \begin{equation}
     \mathcal{H} = \{h\in \mathbb{R}^n : w_i(h)>0, i=1,2,\ldots, n\}.
     \label{eqn:space}
 \end{equation}
 Furthermore, $\nabla u_h$ minimizes the quadratic cost
 \begin{equation}
 \frac{1}{2}\int_{\Omega} \|x - T(x)\|^2d\mu(x)
 \end{equation}
 among all transport maps $T_{\#}\mu=\nu$, where the Dirac measure $\nu = \sum_{i=1}^n \nu_i\delta(y-y_i)$.
 \end{theorem}

The gradient of the above convex energy in Eqn.~\ref{eqn:alexandrov_potential} is given by:
 \begin{equation}
     \label{eqn:gradient}
     \nabla E(h) = (w_1(h) - \nu_1, w_2(h)-\nu_2, \ldots, w_n(h) - \nu_n)^T
 \end{equation}
 The Hessian of the energy is given by
 \begin{equation}
 \frac{\partial w_i}{\partial h_j } = -\frac{\mu(W_i\cap W_j\cap\Omega)}{\|y_i - y_j\|},
 ~\frac{\partial w_i}{\partial h_i } = -\sum_{j\neq i} \frac{\partial w_i}{\partial h_j }
 \label{eqn:Hessian}
 \end{equation}

\begin{figure}[t]
\begin{center}
\begin{tabular}{cc}
\includegraphics[width=0.4\textwidth,angle=90]{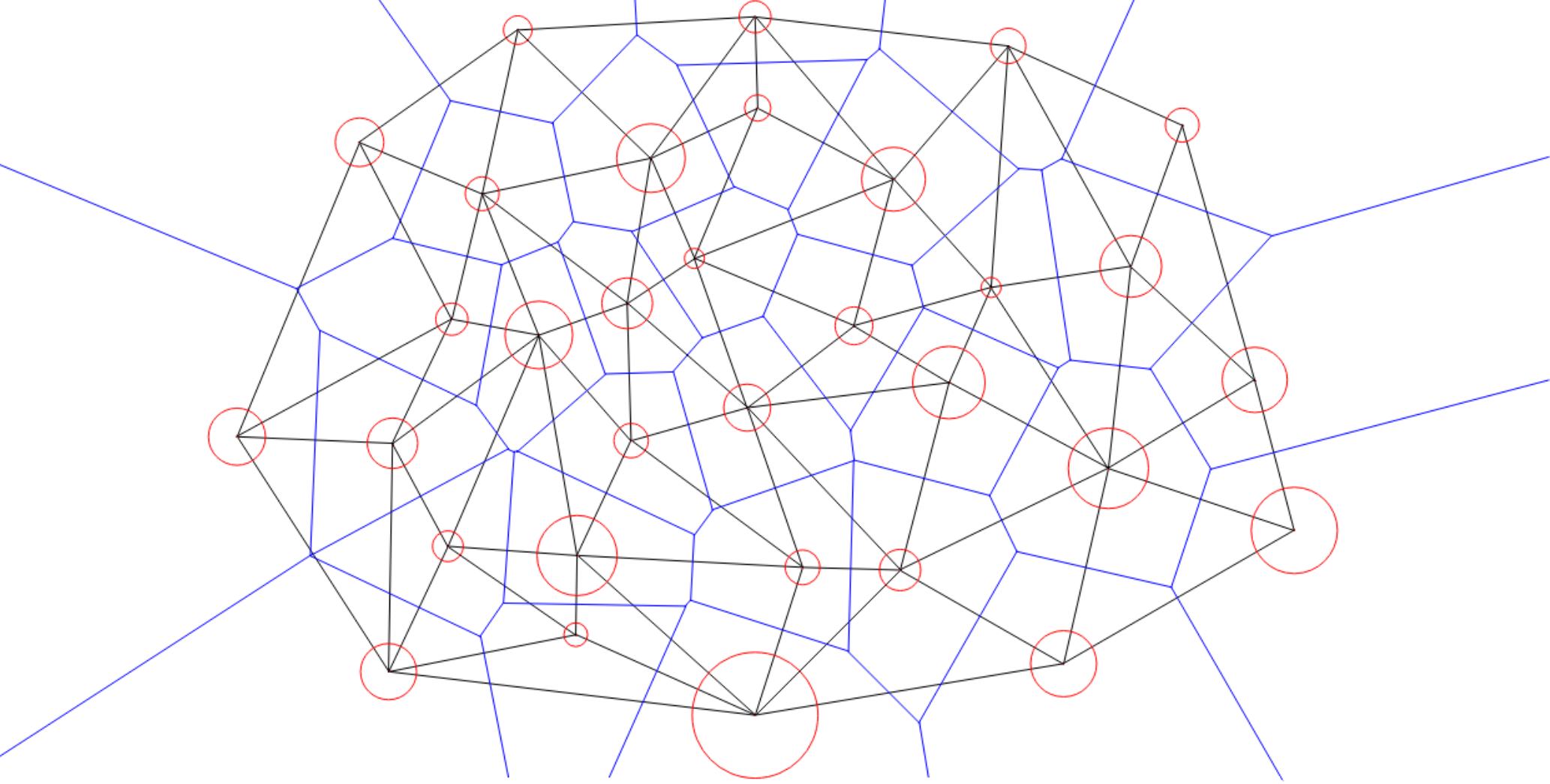} & \!\!
 \includegraphics[width=.45\textwidth]{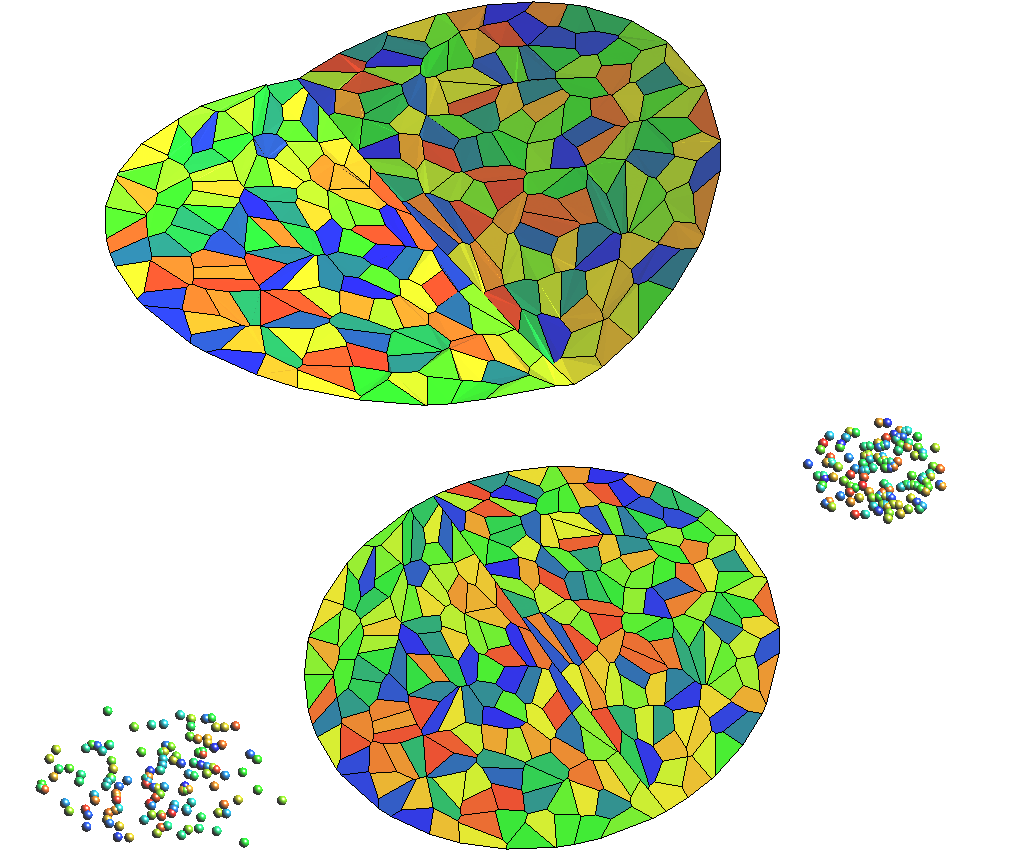}
\end{tabular}
\end{center}\vspace{-5mm}
\caption{The left frame shows Power diagram (blue) and its dual weighted Delaunay triangulation (black). The right frame shows the Optimal transportation map from a disk to two cluster of points.
\label{fig:multi_cluster}}\vspace{-3mm}
\end{figure}

As shown in Fig.~\ref{fig:PL_Brenier_potential}, the Hessian matrix has explicit geometric interpretation. The left frame shows the discrete Brenier potential $u_h$, the right frame shows its Legendre transformation $u_h^*$ using definition \ref{eqn:Legendre_transform}. The Legendre transformation can be constructed geometrically: for each supporting plane $\pi_{h,i}$, we construct the dual point $\pi_{h,i}^*=(y_i,-h_i)$, the convex hull of the dual points $\{\pi_{h,1}^*, \pi_{h,2}^*,\dots, \pi_{h,n}^*\}$ is the graph of the Legendre transformation $u_h^*$. The projection of $u_h^*$ induces a triangulation of $Y=\{y_1,y_2,\dots,y_n\}$, which is the \emph{weighted Delaunay triangulation}. As shown in the left fram of \ref{fig:multi_cluster}, the power diagram in Eqn.\ref{eqn:cell} and weighted Delaunay triangulation are Poincar\`{e} dual to each other: if in the power diagram,  $W_i(h)$ and $W_j(h)$ intersect at a $(d-1)$-dimensional cell , then in the weighted Delaunay triangulation $y_i$ connects with $y_j$. The element of the Hessian matrix Eqn.~\ref{eqn:Hessian} is the ratio between the $\mu$-volume of the $(d-1)$ cell in the power diagram and the length of dual edge in the weighted Delaunay triangulation.

Fig.~\ref{fig:multi_cluster} shows one computational example based on the theorem \ref{thm:gu_yau}.
Suppose the support of the target measure $\nu$ has two connected components, restricted on each component, $\nu$ has a smooth density function. We sample $\nu$ and use a Dirac measure $\hat{\nu}$ to approximate it. By increasing the sampling density, we can construct a sequence $\{\hat{\nu}_k\}$ weakly converges to $\nu$, $\hat{\nu}_k\to \nu$, the Alexandrov solution to $\hat{\nu}_k$ also converges to Alexandrov solution to the Monge-Amp\`{e}re equation with $\nu$.

At each stage, the target $\hat{\nu}_k$ is a Dirac measure with two clusters, the source $\mu$ is the uniform distribution on the unit disk. Each cell on the disk is mapped to a point with the same color. The Brenier potential $u_k$ has a ridge in the middle. Let $k\to \infty$, $u_k\to u$, the ridge on $u_k$ will be preserved on the limit $u$, whose projection is the singularity set $\Sigma_1$ for the limit optimal transportation map $\nabla u$. Along $\Sigma_1$, $\nabla u$ is discontinuous, but the Brenier potential $u$ is always continuous.

\section{Mode Collapse and Regularity}
\label{sec:gan}

Although GANs are powerful for many applications, they have critical drawbacks: first, training of GANs are tricky and sensitive to hyper-parameters, difficult to converge; second, GANs suffer from mode collapsing; third, GANs may generate unrealistic samples. This section focuses on explaining these difficulties using the regularity theorem \ref{thm:regularity} of transportation maps.

\begin{figure}[ht!]
\begin{center}
\begin{tabular}{cc}
 \includegraphics[width=.45\linewidth]{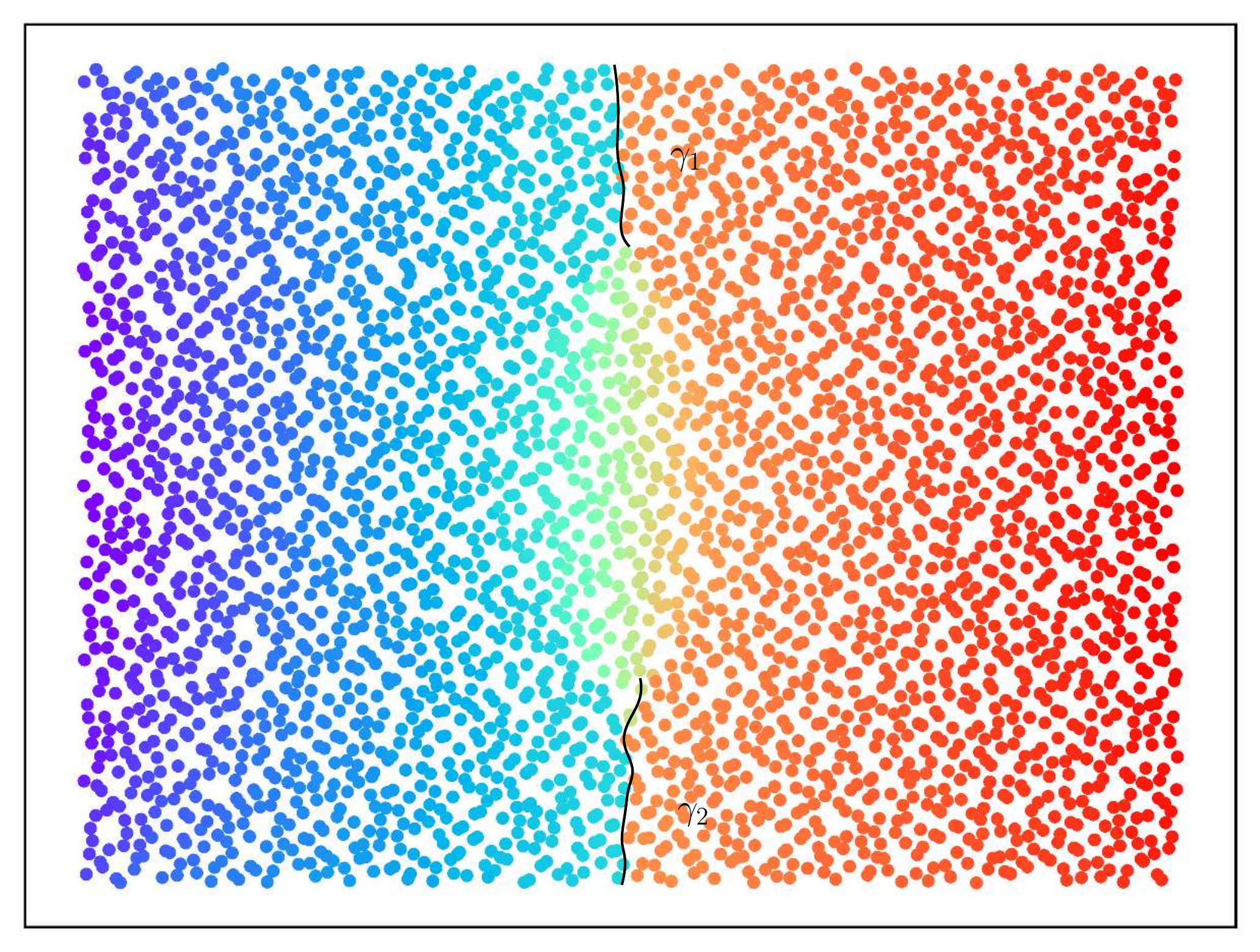}&
 \includegraphics[width=.45\linewidth]{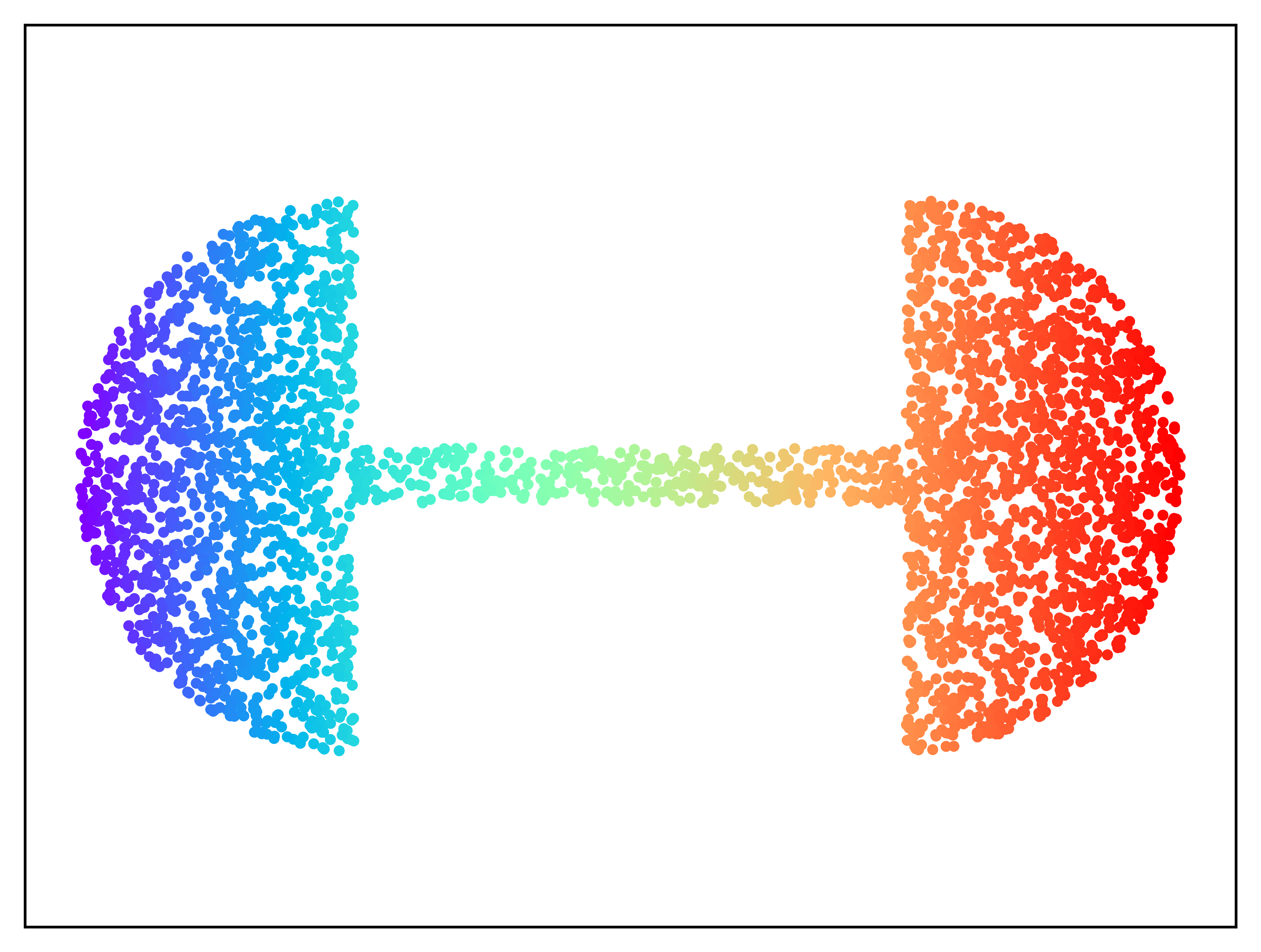}\\
\end{tabular}
\end{center}

\caption{Discontinuous Optimal transportation map, produced by a GPU implementation of algorithm based on theorem \ref{thm:gu_yau}.
\label{fig:dumb_bell}}
\end{figure}

\noindent\textbf{Intrinsic Conflict} 
The difficulty of convergence, mode collapse, and generating unrealistic samples can be explained by the regularity theorem of the optimal transportation map. 

Suppose the support $\Lambda$ of the target measure $\nu$ has multiple connected components, namely $\nu$ has multiple modes, or $\Lambda$ is non-convex, then the optimal transportation map $T:\Omega\to\Lambda$ is discontinuous, the singular set $\Sigma_\Omega$ is non-empty. Fig.~\ref{fig:multi_cluster} shows the multi-cluster case, $\Lambda$ has multiple connected components, where the optimal transportation map $T$ is discontinuous along $\Sigma_1$. Fig.~\ref{fig:dumb_bell} shows even $\Lambda$ is connected , but non-convex. $\Omega$ is a rectangle, $\Lambda$ is a dumbbell shape, the density functions are constants, the optimal transportation map is discontinuous, the singularity set $\Sigma_1 = \gamma_1 \cup \gamma_2$. In general situation, due to the complexity of the real data distributions, and the embedding manifold $\Sigma$, the encoding/decoding maps, the supports of the target measure 
are rarely convex, therefore the transportation mapping can not be continuous globally.

\begin{figure}[ht!]
    \begin{center}
    \begin{tabular}{ccc}
        \includegraphics[width=0.3\textwidth]{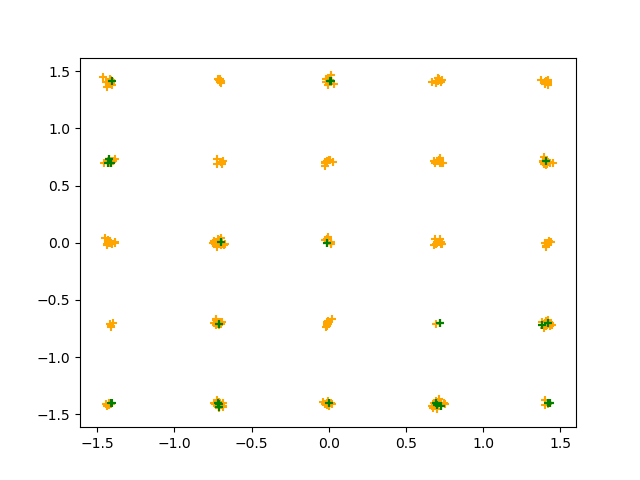}&
        \includegraphics[width=0.3\textwidth]{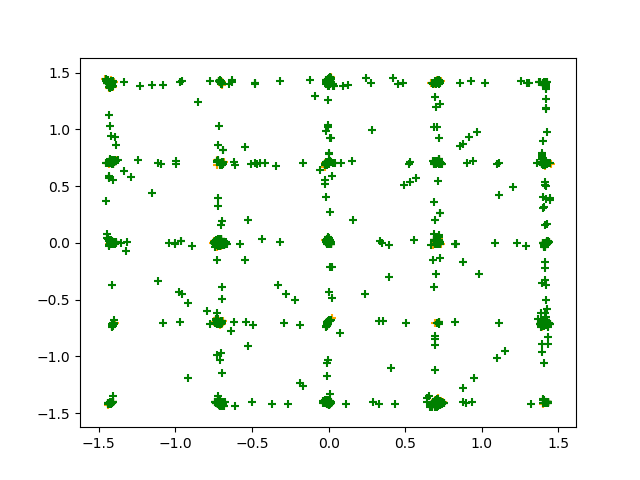}&
        \includegraphics[width=0.3\textwidth]{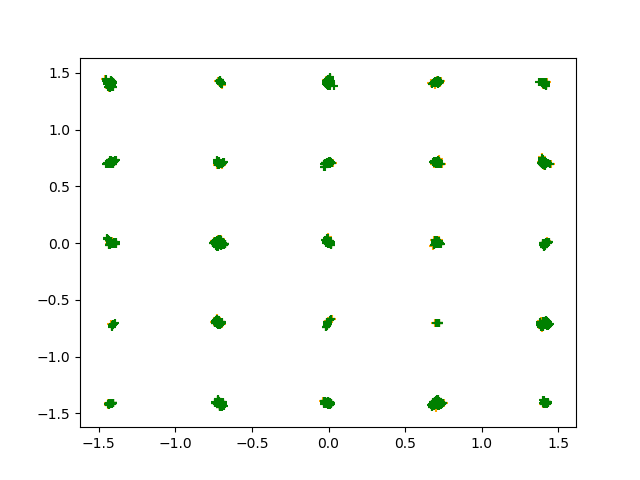}\\
    \end{tabular}
    \end{center}\vspace{-6mm}
    \caption{Comparison between PacGAN and our method to tackle mode collapsing.    \label{fig:comparison}}
\end{figure}

On the other hand, general deep neural networks, e.g. ReLU DNNs, can only approximate continuous mappings. The functional space represented by ReLU DNNs doesn't contain the desired discontinuous transportation mapping. The training process or equivalently the searching process will leads to three situations:
\begin{enumerate}
\item The training process is unstable, and doesn't converge;
\item The searching converges to one of the multiple connected components of $\Lambda$, the mapping converges to one continuous branch of the desired transformation mapping. This means we encounter a mode collapse;
\item The training process leads to a transportation map, which covers all the modes successfully, but also cover the regions outside $\Lambda$. In practice, this will induce the phenomena of generating unrealistic samples. As shown in the middle frame of Fig.~\ref{fig:comparison}.
\end{enumerate} 
Therefore, in theory, it is impossible to approximate optimal transportation maps directly using DNNs.

\noindent\textbf{Proposed Solution} 
The fundamental reason for mode collapse is the conflict between the regularity of the transportation map and the continuous functional space of DNNs. In order to tackle this problem, we propose to compute the Brenier potential itself, instead of its gradient (the transportation map). This is based on the fact that the Brenier potential is always continuous 
under mild conditions, and representable by DNNs, but its gradient is rarely continuous and always outside the functional space of DNNs.

\paragraph{Multi-mode Experiment} We use GPU implementation of the algorithm in Section \ref{sec:alg} to compute the Brenier potential. As shown in Fig.~\ref{fig:comparison}, we compare our method with a recent GAN training method (PacGAN \cite{lin2018pacgan}) that aims to reduce mode collapse. Orange markers are real samples and green markers represent generated ones. Left frame shows a typical case of mode collapse where the generated samples cannot cover all modes. Middle frame shows the result of PacGAN. Although all $25$ modes are captured, the model also generates data that deviates from real samples. Right frame shows the result of our method that precisely captured all modes. It is obvious that our method accurately approximates the target measures and covers all the modes, whereas the method PacGAN generates many fake samples between the modes.

\begin{figure}[ht!]
\begin{center}
\begin{tabular}{cc}
 \includegraphics[width=.4\linewidth]{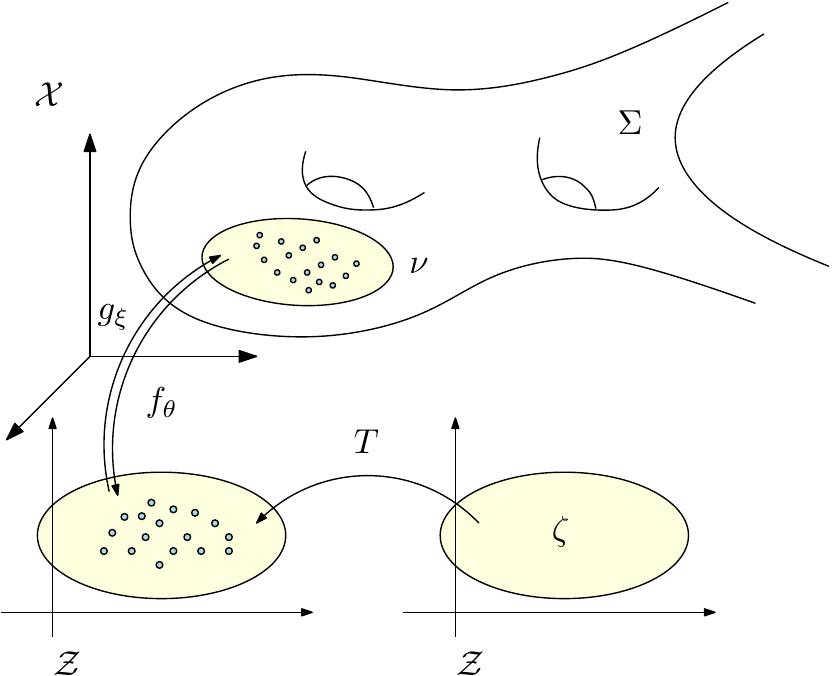}
\end{tabular}
\end{center}\vspace{-5mm}
\caption{AE-OT framework.
\label{fig: AE-OT}}
\end{figure}

\begin{figure}[ht!]
\begin{center}
\begin{tabular}{cc}
 \includegraphics[width=.4\linewidth]{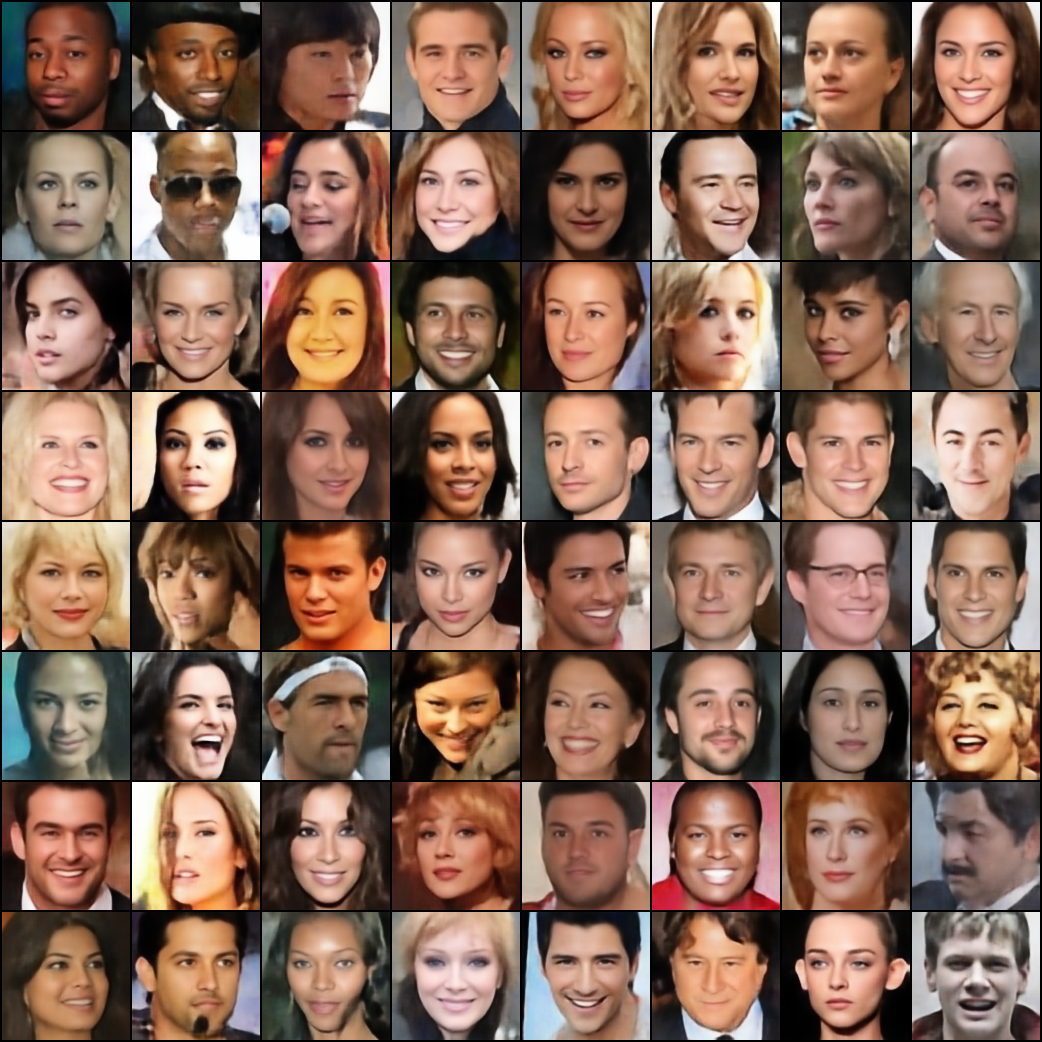}&
 \includegraphics[width=.4\linewidth]{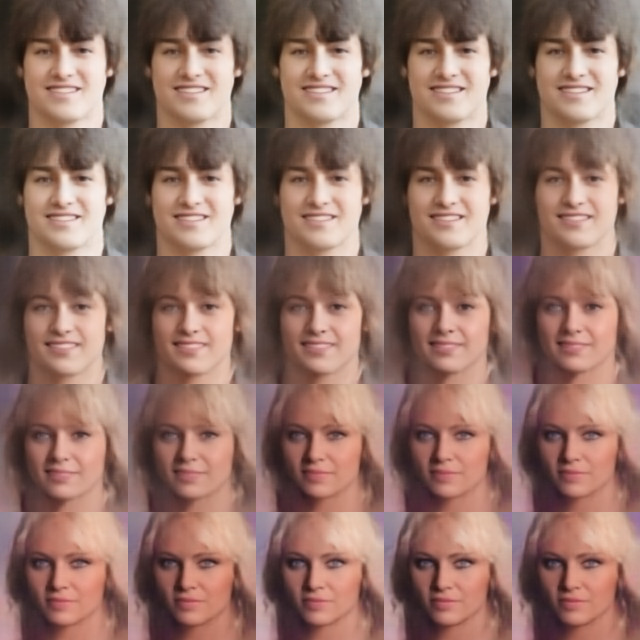}\\
 (a) generated facial images & (b) a path through a singularity.
\end{tabular}
\end{center}\vspace{-5mm}
\caption{Facial images generated by an AE-OT model.
\label{fig:CelebA}}
\end{figure}

\paragraph{Hypothesis Test on CelebA} In this experiment, we want to test our hypothesis: \textbf{In most real applications, the support of the target measure is non-convex, therefore the singularity set is non-empty}.

As shown in Fig.~\ref{fig: AE-OT}, we use an auto-encoder (AE) to compute the encoding/decoding maps from CelebA data set $(\Sigma,\nu)$ to the latent space $\mathcal{Z}$, the encoding map $f_\theta: \Sigma\to \mathcal{Z}$ pushes forward $\nu$ to $(f_\theta)_\# \nu$ on the latent space. In the latent space, we compute the optimal transportation map (OT) based on the algorithm described in Section \ref{sec:alg}, $T: \mathcal{Z}\to\mathcal{Z}$, $T$ maps the uniform distribution in a unit cube $\zeta$ to $(f_\theta)_\# \nu$. Then we randomly draw a sample $z$ from the distribution $\zeta$, and use the decoding map $g_\xi: \mathcal{Z}\to \Sigma$ to map $T(z)$ to a generated human facial image $g_\xi\circ T(z)$. The left frame in Fig.~\ref{fig:CelebA} demonstrates the realist facial images generated by this AE-OT framework.

If the support of the push-forward measure $(f_\theta)_\# \nu$ in the latent space is non-convex, there will be singularity set $\Sigma_k$, $k>0$. We would like to detect the existence of $\Sigma_k$. We randomly draw line segments in the unit cube in the latent space, then densely interpolate along this line segment to generate facial images. As shown in the right frame of Fig.~\ref{fig:CelebA}, we find a line segment $\gamma$, and generate a morphing sequence between a boy with a pair of brown eyes and a girl with a pair of blue eyes. In the middle, we generate a face with one blue eye and one brown eye, which is definitely unrealistic and outside $\Sigma$. This means the line segment $\gamma$ goes through a singularity set $\Sigma_k$, where the transportation map $T$ is discontinuous. \textbf{This also shows our hypothesis is correct, the support of the encoded human facial image measure on the latent space is non-convex. }

As a by-product, we find this AE-OT framework improves the training speed by factor $5$ and increases the convergence stability, since the OT step is a convex optimization. This gives a promising way to improve existing GANs.

\section{Conclusion}
\label{sec:conclusion}

This work builds the connection between the regularity theory of optimal transportation map, Monge-Amp\`{e}re equation and GANs, which gives an theoretic understanding of the major drawbacks of GANs: convergence difficulty and mode collapse.
 
According to the regularity theory of Monge-Amp\`{e}re equation, if the support of the target measure is disconnected or just non-convex, the optimal transportation mapping is discontinuous. General DNNs can only approximate continuous mappings, this intrinsic conflict leads to the convergence difficulty and mode collapse in GANs.

We test our hypothesis that the supports of real data distribution are in general non-convex, therefore the discontinuity is unavoidable using an Autoencoder combined with discrete optimal transportation map (AE-OT framework) on the CelebA data set. The testing result is positive. Furthermore, we propose to approximate the continuous Brenier potential directly based on discrete Brenier theory to tackle mode collapse problem. Comparing with existing methods, this method provides a possible way to achieve higher accuracy and efficiency.

\bibliography{references}
\bibliographystyle{plain}

\pagebreak
\appendix

\section{Kontarovich's Approach}\label{sec:Kontarovichmethod}
Depending on the cost function and the measures, the optimal transportation map between $(X,\mu)$ and $(Y,\nu)$ may not exist. Kontarovich relaxed transportation maps to transportation plans, and defined joint probability measure $\rho: X\times Y\to \mathbb{R}_{\ge 0}$, such that the marginal probability of $\rho$ equations to $\mu$ and $\nu$ respectively.
Formally, let the projection maps be $\pi_x(x,y)=x$, $\pi_y(x,y)=y$, then define mapping class
\begin{equation}
    \Pi(\mu,\nu):=\{\rho:X\times Y\to\mathbb{R}:  (\pi_x)_\#\rho=\mu, (\pi_y)_\#\rho=\nu \}
\end{equation}
\vspace{-8mm}
\begin{problem}[Kontarovich] Given a transport cost function $c: X\times Y\to \mathbb{R}$, find the joint probability measure $\rho:X\to Y$ that minimizes the total transport cost
\begin{equation}
(KP)\hspace{5mm}  \mathcal{W}_c(\mu,\nu) = \min_{\rho \in \Pi(\mu,\nu)} \int_{X\times Y} c(x,y) d\rho(x,y).
    \label{eq:KP}
\end{equation}
\end{problem}
Kontarovich's problem can be solved using linear programming method. Due to the duality of lineary programming, the (KP) Eqn.\ref{eq:KP} can be reformulated as the duality problem (DP) as follows:
\begin{problem}[Duality] Given a transport cost function $c: X\times Y\to \mathbb{R}$, find the function $\varphi:X\to \mathbb{R}$ and $\psi:Y\to \mathbb{R}$, such that
\begin{equation}
(DP)\hspace{5mm}  \max_{\varphi,\psi} \left\{\int_X \varphi(x) d\mu + \int_Y \psi(y) d\nu~:~ \varphi(x)+\psi(y)\le c(x,y) \right\}
    \label{eqn:DP}
\end{equation}
\end{problem}
The maximum value of Eqn.\ref{eqn:DP} gives the Wasserstein distance. Most existing Wasserstein GAN models are based on the duality formulation under the $L^1$ cost function.
\begin{definition}[$c$-tranformation]
The \emph{$c$-tranformation} of $\varphi:X\to\mathbb{R}$ as $\varphi^c:Y\to \mathbb{R}$:
\begin{equation}
    \varphi^c(y) = \inf_{x\in X}( c(x,y) - \varphi(x)).
\end{equation}
\end{definition}
Then the duality problem can be rewritten as
\begin{equation}
(DP)\hspace{5mm}  \mathcal{W}_c(\mu,\nu)=\max_\varphi
\int_X \varphi(x) d\mu + \int_Y \varphi^c(y) d\nu,
    \label{eqn:DP2}
\end{equation}
where $\varphi$ is called the \emph{Kontarovich's potential}.

\end{document}